\documentclass{article}
\PassOptionsToPackage{numbers, compress}{natbib}

\usepackage[dblblindworkshop, final]{neurips_2025}

\usepackage[utf8]{inputenc} 
\usepackage[T1]{fontenc}    
\usepackage[
    hidelinks,
    colorlinks=black,
    linkcolor=black,
    urlcolor=blue,
    citecolor=black
]{hyperref}       
\usepackage{url}            
\usepackage{booktabs}       
\usepackage{amsfonts}       
\usepackage{nicefrac}       
\usepackage{microtype}      
\usepackage{xcolor}
\usepackage{graphicx}
\usepackage{amsmath}
\usepackage{pifont}
\usepackage{subcaption}
\usepackage{booktabs} 
\usepackage{tabularx} 
\usepackage{multirow} 
\usepackage{tcolorbox}
\usepackage{scrextend}
\usepackage{placeins}
\usepackage{nord}

\usepackage{tcolorbox}
\tcbuselibrary{skins,breakable}

\newtcolorbox[auto counter,
  number within=section,
  list inside=prompt,
  list type=prompt
]{llmprompt}[2][]{
  float, 
  title={Prompt \thetcbcounter: #2}, 
  colback=nord5,       
  colframe=nord1,     
  fontupper=\ttfamily,      
  boxrule=1pt,              
  arc=3mm,                  
  boxsep=5pt,               
  left=5pt, right=5pt, top=2pt, bottom=2pt,
  verbatim,                 
  #1 
}

\title{Are We Asking the Right Questions? On Ambiguity in Natural Language Queries for Tabular Data Analysis}
\workshoptitle{AI for Tabular Data}

\author{%
  \textbf{Daniel Gomm}$^{1,2}$ \quad Cornelius Wolff$^{1,2}$ \quad Madelon Hulsebos$^1$ \\
  $^1$Centrum Wiskunde \& Informatica \quad $^2$University of Amsterdam\\
  \texttt{\{daniel.gomm, cornelius.wolff, madelon.hulsebos\}@cwi.nl} \\
}

\makeatletter
\renewcommand{\@noticestring}{Preprint. Accepted to the AI for Tabular Data workshop at EurIPS 2025}
\makeatother

\begin{document}

\maketitle

\vspace{-0.25cm}

\begin{abstract}

Natural language interfaces to tabular data must handle ambiguities inherent to queries. Instead of treating ambiguity as a deficiency, we reframe it as a feature of \textit{cooperative interaction} where users are intentional about the degree to which they specify queries. We develop a principled framework based on a shared responsibility of query specification between user and system, distinguishing unambiguous and ambiguous cooperative queries, which systems can resolve through reasonable inference, from uncooperative queries that cannot be resolved. Applying the framework to evaluations for tabular question answering and analysis, we analyze queries in 15 datasets, and observe an uncontrolled mixing of query types neither adequate for evaluating a system's accuracy nor for evaluating interpretation capabilities. This conceptualization around cooperation in resolving queries informs how to design and evaluate natural language interfaces for tabular data analysis, for which we distill concrete directions for future research and broader implications.

\end{abstract}

\vspace{-0.15cm}

\section{Rethinking Ambiguity in Analytical Natural Language Queries}
\label{sec:introduction}
Natural language interactions with data systems are characterized by ambiguities, as indicated by studies of real-world text-to-SQL systems, which find that a large share of queries is ambiguous \cite{wangKnowWhatDont2023}. 
These ambiguities are amplified in open-domain tabular analysis. In this setting, users state an \textit{insight need} in natural language that requires retrieving, transforming, and executing analysis over tabular data. Unlike text-to-SQL over fixed databases \cite{yuSpiderLargeScaleHumanLabeled2018, NEURIPS2023_83fc8fab}, this setting provides minimal contextual scaffolding, requiring that the query specifies the analytical procedure and the data to which it is applied, so that a system can identify and analyze the relevant data to satisfy the insight need.

If users had a perfect understanding of the underlying data and analysis, they could provide \textit{a \underline{platonic query} which cannot be interpreted in any other way than mapping to the necessary and sufficient set of relevant data items and a fully parameterized analytical methodology appropriate to address the query intent.} 
In practice, such perfect specification is unrealistic. Users express intent naturally, often leaving details implicit. Indeed, Belkin's Anomalous States of Knowledge hypothesis~\cite{belkin_anomalous_1980} established that users inherently cannot specify precisely what they need, because the very information need stems from their incomplete knowledge. Yet the prevailing response in tabular data analysis is to treat this ambiguity as a technical deficiency, focusing on post-hoc detection, classification and resolution to uncover a single, latent user intent \cite{yuSpiderLargeScaleHumanLabeled2018, NEURIPS2024_a4c942a8, wangKnowWhatDont2023, gommMetadataMattersDense2025, dingAmbiSQLInteractiveAmbiguity2025, saparinaDisambiguateFirstParse2025}, instead of conceptualizing why and how users (under)specify their queries.



We argue for a shift in perspective: Instead of a problem to be fixed, ambiguity in user-system cooperation is an expression of a user's understanding of language, and a signal about user intent and implicit reliance on division of labor. Based on this framing, we develop (1) a framework to characterize analytical queries through the lens of cooperative interaction, defining them by what a user makes explicit versus what a system must infer. From the framework, we derive (2) evaluation criteria and apply them to 15 benchmarks for natural language interfaces to tabular data, revealing systematic misalignments with open-domain requirements. Finally, we distill (3) broader implications for better design and evaluation of tabular data analysis systems.

\section{Analytical Queries through the Lens of Cooperative Interaction}
\label{sec:framework}


Users interact with tabular analysis systems by expressing their insight needs and analytical intentions in natural language queries. These queries are shaped by a user's naturally evolving, often incomplete, internal mental models of the system and the underlying data, acting as their interaction partner \cite{normanObservationsMentalModels1983}. Their understanding of system and data, guides how users formulate their insight needs. 

Since queries unfold as natural language interactions, we analyze them through the lens of cooperative communication. Participants in communication coordinate on mutual understanding through a collaborative process governed by least collaborative effort~\cite{clarkGroundingCommunication1991, clarkReferringCollaborativeProcess1986}, where Grice's cooperative principle~\cite{griceLogicCoversation1975} specifies that they provide sufficient but not excessive information (maxim of quantity), while being truthful (maxim of quality). These cooperative principles have been applied broadly to NLP systems~\cite{krauseGriceanMaximsNLP2024} and empirically validated in human-AI interaction~\cite{panfiliHumanAIInteractionsGricean2021}. Here, we apply them specifically to characterize how users specify analytical queries over tabular data.

\begin{figure}[ht]
    \centering
    \includegraphics[width=\linewidth]{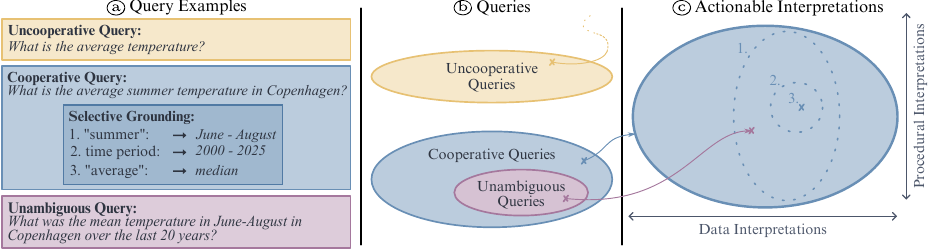}
    \caption{\textcircled{a}: Queries with different levels of specification. \textcircled{b} \& \textcircled{c}: Relationship of queries and actionable interpretations. \textcircled{c} shows iterative selective grounding of the cooperative query in \textcircled{a}, progressively narrowing the set of possible interpretations until arriving at a singular interpretation.
    }
    \label{fig:cooperative_queries_extended_v2}
\end{figure}

Consider the query ``What is the average summer temperature in Copenhagen?'' in Figure \ref{fig:cooperative_queries_extended_v2} \textcircled{a}. To respond, a system must determine both what analytical procedure to perform (e.g., mean, median) and what data to apply it to (which time period; what is ``summer''). \textit{We term the unique combination of a specific analytical procedure and the exact data to which it is applied as \underline{actionable query interpretation}.} Deriving actionable interpretations requires grounding both the analytical intent (the operation and its parameters) and the data scope (entities, temporal boundaries, and domain constraints; see Appendix \ref{sec:appendix:query_specification_framework} for details) into concrete analytical operations and data selections. This grounding succeeds through user-system cooperation, where both parties adhere to shared expectations: users provide information they deem necessary, systems can infer what remains implicit, with both parties avoiding unnecessary ambiguity. Based on this cooperative framing we posit that \textit{a \underline{cooperative query} expresses an insight need such that it provides sufficient specification, either explicitly or through reasonable inference, to identify at least one valid, actionable query interpretation.} 
Conversely, an \textit{\underline{uncooperative query}} is underspecified to a degree that creates \textit{irresolvable ambiguity}, providing insufficient basis for the system to identify a valid, actionable interpretation, as depicted in Figure \ref{fig:cooperative_queries_extended_v2} \textcircled{b} \& \textcircled{c}.
This ambiguity may result from a misaligned mental model of the system's capabilities and the context of the interaction, leading users to provide insufficient information or express analytical intent too vaguely, effectively violating the maxim of quantity. For instance, the query ``What is the average temperature?'' omits critical information about the location, which has no strong convention or well-defined set of reasonable interpretations, rendering the system unable to make a well-founded grounding decision.

Deriving an actionable interpretation from a cooperative query requires \textit{grounding} all its semantic components, both the analytical procedure to be performed and the data it applies to. 
Drawing from the maxim of quantity, we observe that users naturally omit information they expect a cooperative system to infer. This creates a division of labor, where grounding is achieved through two mechanisms: \ding{172} \textit{user-provided grounding}, where information is furnished directly within the query, either explicitly (e.g., ``Apple Inc.'') or contextually (e.g., ``past 20 years,'' given the current year is known), and \ding{173} \textit{system-inferred grounding}, where a cooperative user delegates grounding to the system. 
This delegation is successful under either of two communicative purposes. First, \textit{conventional grounding}, where ambiguity is resolved by applying a near-universal convention or piece of common-sense knowledge (e.g., ``highest mountain'' \ding{221} ``in the world''). Second, \textit{selective grounding}, where users deliberately leave choices underspecified to grant the system agency in selecting from well-defined reasonable options (e.g., ``relationship between...'' \ding{221} Pearson/Spearman correlation, other measure).\\
{\color{nord10_darker} \ding{222} \textbf{Underspecification is a feature, not a problem.} Understanding ambiguity as intentional division of labor that gives systems agency over data selection and methodology fundamentally shifts the view of what systems should do, and thus how we should approach systems design and evaluation.}

Cooperative queries vary in how much selective grounding they require to derive an actionable interpretation. At the end of this spectrum lie queries requiring no selective grounding at all. \textit{Cooperative queries that map to a singular actionable interpretation through user-provided and conventional grounding alone are  \underline{unambiguous queries}.} As shown in Figure \ref{fig:cooperative_queries_extended_v2} \textcircled{a}, the unambiguous query example leaves no room for discretionary system choices in analytical procedure or data scope.\\
{\color{nord10_darker} \ding{222} \textbf{The more complex an analytical query, the more grounding work is required.} While unambiguous queries are easily formulated for simple lookup or aggregations, the specification burden increases substantially in diagnostic analysis, predictive modeling, or prescriptive recommendations.} 

The cooperative query framework generalizes broadly beyond open-domain tabular analysis to any natural language interface over tabular data. For instance, in text-to-SQL systems users know with which database they interact.
This confined data environment itself provides significant context for the cooperation, requiring less explicit grounding of what data users refer to. Thus, the framework's application depends on the user's mental model. Regardless of the task, the fundamental principles of distinguishing between (un-)cooperative queries, understanding the division of grounding labor, and recognizing when selective grounding is appropriate, remain central to system design and evaluation.


\section{What Constitutes a Useful Analytical Query for Evaluation?}
\label{sec:analysis}
The framework of cooperative queries has direct implications for evaluating open-domain tabular analysis systems. Performance is measured by a system's ability to correctly satisfy a user's insight need, but the framework reveals that the ``correctness`` of a response fundamentally depends on the query's specification and interpretation. Sound evaluation should thus isolate a systems interpretation capabilities from execution accuracy. These distinct evaluation purposes are served by different types of cooperative queries. Unambiguous queries map to a single actionable interpretation, making them uniquely suited for assessing execution accuracy against a gold solution without confounding variables from ambiguity in interpretations. Conversely, cooperative queries that delegate analytical decisions to the system through selective grounding can be used to evaluate a system's complementary ability to make reasonable, human-aligned choices when faced with controlled ambiguity. Both query types also assess how systems apply conventional grounding to resolve references through common-sense knowledge. Finally, uncooperative queries lack veryfiable interpretations by definition, making them unsuitable for evaluating execution performance. Instead, they can be used to test a systems robustness to unanswerable queries.\\
{\color{nord10_darker} \ding{222} \textbf{Evaluate with queries fit for the evaluation objective.} Queries should support the evaluation objective. Evaluating execution accuracy against ground truth solutions requires unambiguous queries; evaluating interpretation capabilities requires cooperative queries; uncooperative queries can be used to evaluate robustness.}



Beyond the semantic interpretability of queries, they should also be realistically formulated to reflect authentic user interactions. In the open-domain setting, a user's mental model does not encompass exact knowledge of underlying data structures or contents \cite{voorheesTRECQuestionAnswering2001}. Consequently, authentic \textit{\underline{data-independent queries} are formulated independently from specific dataset characteristics.} Yet, benchmarking datasets are often constructed around specific tables by annotators \cite{nanFeTaQAFreeformTable2022, chenOpenQuestionAnswering2020, huangDACodeAgentData2024}, synthetically \cite{huInfiAgentDABenchEvaluatingAgents2024, wuMMQAEvaluatingLLMs2024}, or in a hybrid manner \cite{wuTableBenchComprehensiveComplex2025, zhangCRTQADatasetComplex2023}. This can lead to leakage of privileged information into queries. \textit{We term queries that use knowledge inaccessible to users in the open-domain setting as \underline{data-privileged queries}.} This manifests in direct references to structural elements like column headers (e.g., \textit{"index"}, \textit{"first\_name"}), specific values not in the public domain (e.g., \textit{"order \#A729-T"}, \textit{"user that ordered pizza for 102.07\$"}), or even the data containers themselves (e.g., \textit{"the country information table"}). Such references provide a strong but unrealistic signal linking the query to specific data structures or contents, fundamentally undermining the open-domain premise.

To assess ambiguity and data-privilege in existing benchmarks, we analyze a broad range of 15 datasets spanning tabular question answering \cite{pasupatCompositionalSemanticParsing2015, luDynamicPromptLearning2022, zhangCRTQADatasetComplex2023, chengHiTabHierarchicalTable2022, kweonOpenWikiTableDatasetOpen2023, chenOpenQuestionAnswering2020, nanFeTaQAFreeformTable2022, wuTableBenchComprehensiveComplex2025, zhaoMultiHierttNumericalReasoning2022,wuMMQAEvaluatingLLMs2024}, text-to-SQL \cite{yuSpiderLargeScaleHumanLabeled2018, NEURIPS2023_83fc8fab}, and data analysis \cite{huangDACodeAgentData2024, laiKramaBenchBenchmarkAI2025, huInfiAgentDABenchEvaluatingAgents2024}.
While not all designed for the full end-to-end scope of open-domain tabular analysis, these datasets are commonly used to evaluate systems in this area \cite{chenOpenQuestionAnswering2020, nanFeTaQAFreeformTable2022, kongOpenTabAdvancingLarge2023,ji2025target, zhangMURREMultiHopTable2025, laiKramaBenchBenchmarkAI2025}. We characterize queries in each benchmark along their data-independence and cooperative interpretability using LLM-based classifiers validated against expert annotations, with the LLM-judge achieving agreement at or above human inter-annotator level across all dimensions (full setup, prompts, and validation in Appendix \ref{sec:appendix:query_classification_methodology}).

\begin{figure}[t]
\centering
\begin{subfigure}{.48\textwidth}
  \centering
    \includegraphics[width=\textwidth]{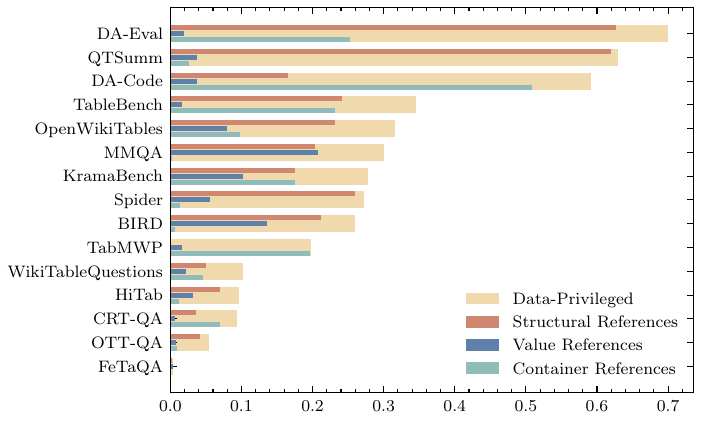}
    \caption{Share of data-privileged queries in datasets.}
    \label{fig:data_privileged_queries}
\end{subfigure}%
\begin{subfigure}{.48\textwidth}
  \centering
    \includegraphics[width=\textwidth]{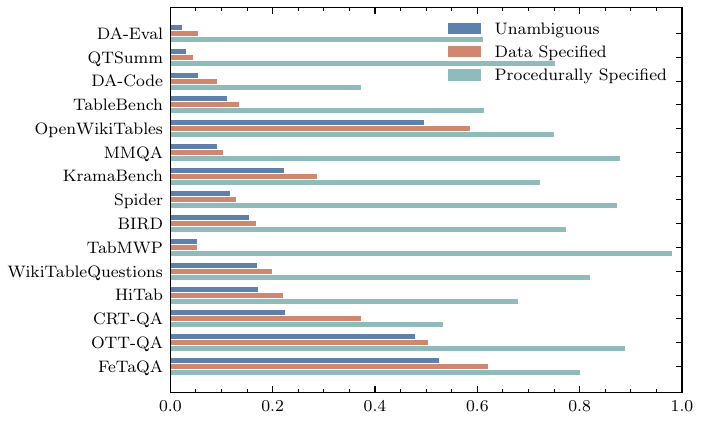}
    \caption{Share of unambiguous queries in datasets.}
    \label{fig:query_specification}
\end{subfigure}
\caption{Analysis of query characteristics across 15 tabular benchmarks.}
\label{fig:analysis_results}
\vspace{-0.3cm}
\end{figure}

Our analysis in Figure \ref{fig:analysis_results} reveals systematic issues in existing benchmarks. First, Figure \ref{fig:data_privileged_queries} shows that many datasets, especially for complex tabular analysis, are saturated with data-privileged queries, providing shortcuts that are unavailable in a true open-domain setting. 
Second, Figure \ref{fig:query_specification} shows that unambiguous queries that have a single, verifiable interpretation are a small fraction across datasets. 
Thus, only few queries are suitable for testing pure execution accuracy, while the remaining queries are not suitable for evaluating interpretative capabilities either since the datasets admit only a single intended answer.
The share of unambiguous queries is particularly low in complex tabular analysis benchmarks like DA-Eval \cite{huInfiAgentDABenchEvaluatingAgents2024} and DA-Code \cite{huangDACodeAgentData2024}, reflecting the inherent challenges in fully specifying complex queries.
While many queries are procedurally specified, they remain ambiguous due to an underspecified data scope, requiring either selective grounding or rendering them irresolvable.\\
{\color{nord10_darker} \ding{222} \textbf{Existing evaluations fail to isolate system capabilities.} Benchmarks mix unambiguous, cooperative, and uncooperative queries without distinction. Evaluating without differentiating conflates a system's execution accuracy with its cooperative ability to make human-aligned grounding decisions.}

\vspace{-0.1cm}

\section{Elevating Natural Language Systems for Tabular Data Analysis}
\label{sec:discussion}
\vspace{-0.2cm}
By understanding ambiguity and grounding characteristics through the lens of cooperative queries, we envision more targeted evaluations and systems that are better aligned with authentic user interactions.

\textbf{Towards more effective evaluation practices.}
Existing datasets can be immediately improved by augmenting them with annotations of query specification levels and grounding requirements. This enables stratified evaluations by query type, diagnosing failures by distinguishing execution accuracy on unambiguous queries from selective grounding on cooperative queries.


\textbf{Novel datasets for iterative query refinement.}
Datasets for iterative query refinement should provide multiple grounding paths from underspecified queries to actionable interpretations, systematically exploring how systems handle controlled ambiguity.
This allows testing both a system's ability to recognize when selective grounding is needed and its capacity to make human-aligned choices or appropriately request clarification. Critically, evaluating these capabilities requires a shift to flexible evaluation protocols that assess the validity and alignment of interpretations rather than enforcing a single ``correct'' answer. While recent works start exploring this direction by testing against sets of pre-defined solutions \cite{laiKramaBenchBenchmarkAI2025,guBLADEBenchmarkingLanguage2024}, they do not account for the full range of reasonable grounding choices. 

\textbf{Designing systems for cooperative interaction.}
Designing systems around cooperation with users enables a productive division of grounding labor. This empowers systems to take a more proactive role in interpreting queries within their boundaries, dynamically grounding them in the context of available data and appropriate analytical procedures. Recognizing the significance of these interpretations, systems should disclose their grounding choices, enabling users to intervene on misalignments. 

\textbf{From single-shot to cooperative dialog.} While we have focused on single-shot query execution, irresolvable ambiguities should be handled through mixed-initiative specification~\cite{horvitz_principles_1999, allen_mixedinitiative_1999}, cooperatively engaging the user to resolve ambiguities rather than fail or guess. A key challenge lies in detecting and balancing when to automate grounding decisions and when to require explicit user consultation.

\textbf{Towards open-domain tabular data analysis.}
We are still in the early stages of end-to-end open-domain tabular data analysis systems, with only preliminary research on table retrieval and analytical tool selection for complex analytical workloads. The path forward requires a shift towards holistically integrated systems, grounded in the cooperative understanding of interactions discussed here, to effectively navigate the full workflow translating insight needs to value.

\section*{Acknowledgements}
This work was supported by the Dutch Research Council (NWO, grant NGF.1607.22.045) and a grant by SAP. We also gratefully acknowledge support from OpenAI and Google through research credits.

\bibliography{references}


\newpage
\appendix
\section*{Appendix}
\section{Conceptualizing types of Query Specification}
\label{sec:appendix:query_specification_framework}

As discussed in Section \ref{sec:framework}, deriving actionable interpretations from queries requires grounding both the analytical procedure and the data scope. To systematically characterize queries, we introduce a framework that decomposes specification requirements across two primary axes: \textit{procedural specification} and \textit{data specification}. Each axis comprises multiple dimensions that capture distinct aspects of how users express their information needs.

\textbf{Procedural Specification} captures how clearly a query defines the analytical operations to be performed. This encompasses two dimensions: \textit{Intent Specification} evaluates whether the query expresses a clear insight need through an analytical goal, distinguishing interpretable insight needs that map to executable operations from vague requests like "tell me about" or "insights on." \textit{Methodological Specification} evaluates whether the analytical methods, parameters, and calculations needed to execute the intent are sufficiently defined. This includes aggregation functions (mean vs. median), correlation methods (Pearson vs. Spearman), ranking metrics, and other analytical parameters. Together, these dimensions characterize the analytical procedure component of actionable interpretations.

\textbf{Data Specification} captures how clearly a query defines what data to analyze and how to structure the analysis across that data. We distinguish between specifications that define the scope of data entities (adjectival role) and those that constrain the analytical structure (adverbial role). The framework comprises five dimensions organized into two layers:

\textit{Entity Scope Specification} defines which data subjects are being analyzed through three sub-dimensions: \textit{Core Entity Specification} identifies the fundamental analytical subject (e.g., "revenue," "hospitals," "employees"). \textit{Entity Temporal Bounds} specifies when these entities exist or occur (e.g., "2024 Olympics," "Q4 revenue"), answering "which instances of this entity?" \textit{Entity Domain Bounds} specifies where or in what context these entities exist (e.g., "California hospitals," "European universities"), further constraining entity scope.

\textit{Analytical Constraint Specification} defines how the analysis should be structured across the data through two sub-dimensions: \textit{Temporal Analytical Structure} specifies whether and how the analysis should be organized temporally (e.g., "by year," "over time," "year-over-year comparison"), answering "how should time structure this analysis?" \textit{Domain Analytical Structure} specifies whether and how the analysis should be organized across spatial or conceptual domains (e.g., "by region," "across countries," "per department"), answering "how should space/context structure this analysis?"

This distinction between entity-bounding (adjectival) and analysis-structuring (adverbial) roles prevents ambiguous attribution: "California hospitals in 2024" uses temporal and domain specifications to bound entity scope, while "compare hospitals by state over years" uses them to structure analytical comparisons.

\textbf{Relationship to Query Cooperativeness} This specification framework directly operationalizes the cooperative query concepts from the cooperative query framework introduced in Section \ref{sec:framework}. Uncooperative queries contain at least one \textit{irresolvable dimension}, lacking sufficient information for the system to derive any valid actionable interpretation. This framework thus provides an orthogonal component to the cooperative query framework, enabling a principled application of the framework along a fixed set of dimensions.

\section{Overview of Analyzed Datasets}
\label{sec:appendix:datasets_overview}

Our analysis examines the 15 datasets presented in Table \ref{tab:dataset_comparison}, spanning tabular question answering, text-to-SQL, and data analysis tasks. These datasets draw from diverse data sources including Wikipedia, statistical reports, company financials, scientific publications, and curated databases from platforms like Kaggle and GitHub. Query complexity ranges from simple factual retrieval to sophisticated analytical tasks such as trend analysis, correlation, regression, and classification. While diverse in these regards, the datasets mostly expect a singular "gold" interpretation, leaving no room for competing, valid query interpretations.

\begin{table}[htbp]
\centering
\caption{Overview of the analyzed datasets and their characteristics.}
\label{tab:dataset_comparison}
\footnotesize 
\begin{tabularx}{\textwidth}{p{2cm} r p{2.5cm} X p{2.5cm}}
\toprule
\textbf{Dataset} & \textbf{\#Queries} & \textbf{Table Sources} & \textbf{Query types} & \textbf{Expected Outputs} \\
\midrule
WikiTable-Questions \cite{pasupatCompositionalSemanticParsing2015} & 14,151 & Wikipedia & Factual retrieval & Exact Value \\
TabMWP \cite{luDynamicPromptLearning2022}  & 38,901 & Wikipedia & Factual retrieval, Aggregation & Free Form Text, Multiple Choice \\
CRT-QA \cite{zhangCRTQADatasetComplex2023} & 728 & Wikipedia & Factual retrieval, Aggregation, Comparative & Exact Value \\
HiTab \cite{chengHiTabHierarchicalTable2022} & 10,672 & Wikipedia, Statistical reports & Factual retrieval, Aggregation, Comparative & Exact Value \\
OpenWiki-Table \cite{kweonOpenWikiTableDatasetOpen2023} & 67,023 & Wikipedia & Factual Retrieval, Aggregation, Comparative & Exact Value \\
OTT-QA \cite{chenOpenQuestionAnswering2020} & 4,372 & Wikipedia & Factual Retrieval, Aggregation, Comparative & Free Form Text \\
FeTaQA \cite{nanFeTaQAFreeformTable2022} & 10,330 & Wikipedia & Factual Retrieval, Aggregation, Comparative & Free Form Text \\
TableBench \cite{wuTableBenchComprehensiveComplex2025} & 886 & Wikipedia & Factual Retrieval, Aggregation, Trend Analysis & Exact Value \\
QTSumm \cite{zhaoMultiHierttNumericalReasoning2022} & 10,440 & Annual Company Reports & Factual retrieval, Aggregation & Exact Value \\
MMQA \cite{wuMMQAEvaluatingLLMs2024} & 3,313 & Wikipedia & Factual Retrieval, Aggregation, Comparative & Exact Structured Output \\
Spider \cite{yuSpiderLargeScaleHumanLabeled2018} & 11,840 & Example databases from courses & Factual retrieval, Aggregation & SQL, Execution Result \\
BIRD \cite{NEURIPS2023_83fc8fab} & 10,962 & Kaggle, CTU Prague, custom & Factual retrieval, Aggregation, Comparative & SQL, Execution Result \\
DA-Code \cite{huangDACodeAgentData2024} & 500 & Github, Kaggle, Web & Aggregation, Characterization, Comparative, Trend Analysis, Correlational, Regression, Classification & Structured Text Output, Chart, Table \\
KramaBench \cite{laiKramaBenchBenchmarkAI2025} & 104 & Scientific Works & Aggregation, Comparative, Trend Analysis, Regression & Python \\
DA-Eval \cite{huInfiAgentDABenchEvaluatingAgents2024} & 257 & Github & Aggregation, Characterization, Correlational, Regression, Classification & Exact Value \\
\bottomrule
\end{tabularx}
\end{table}

\section{Evaluation Setup for Classifying Queries}
\label{sec:appendix:query_classification_methodology}

To analyze the characteristics of queries across the 15 benchmarks (see Appendix \ref{sec:appendix:datasets_overview}), we randomly sampled 500 queries from each dataset. We then employed LLM-based classifiers to systematically assess each query's properties. The methodology for each classification task is detailed below. You can find the code, prompts, and data in this repository:\\ \url{https://github.com/trl-lab/open-domain-query-classification}.

\subsection{Data-Independence}

To identify data-privileged queries, we developed an LLM-based classifier to assess data-independence along three dimensions: Structural References, Value References, and Container References. We employed gpt-5-mini-2025-08-07 as the LLM-judge using Prompt \ref{prompt:data_privilege}. To ensure robust and stable classifications, we used a self-consistency method, sampling five classifications for each query and assigning the final label based on a majority vote (minimum three identical votes). For the final analysis presented in Figure \ref{fig:data_privileged_queries}, we aggregated the "Obscure" and "True" labels for Value References into a single "False" (data-independent) category.

\begin{llmprompt}{Data-Independence Classification Prompt}
\label{prompt:data_privilege}
\tiny
You are an expert data analyst specializing in human-computer interaction and natural language interfaces for databases. Your task is to analyze a user query and determine if it is schema-dependent.\\
\\
A query is schema-dependent if its phrasing suggests the user has prior knowledge of the underlying data's specific structure or content. In a true open-domain setting, a user would ask a natural, conceptual question about the world, not a question crafted around specific datasets they have already seen. In contrast, a schema-independent query is a natural, conceptual question about the world. Analytical complexity is NOT a sign of schema-dependence. A complex question with many conditions can still be perfectly schema-independent.\\
\\
You will classify each query along three dimensions of schema-dependence. Analyse and classify each query along the definition of dimensions below. Keep the analysis short if the classification is clear.\\
\\
Classification Dimensions\\

- Structural Reference (structural\_reference: bool):
\begin{addmargin}[10pt]{0pt}
    - Set to true if: The query's terminology, especially for nouns and attributes, sounds more like a database column header than a natural, conceptual question. The language feels copied or adapted from a specific schema. The core question is: "Is it more likely the user copied this exact term from a table's column headers, or that they independently came up with this phrasing?"\\
    - What to look for:\\
    \begin{addmargin}[10pt]{0pt}
        - Code-like Syntax: Obvious formatting like snake\_case or camelCase (e.g., asking for the SalePrice).\\
        - Database-Specific Concepts: The use of words that are common in data management but not in everyday questions (e.g., asking for a record's id, key, or index).\\
        - Unnatural Phrasing: Unusual compound terms, hyphenated words, or overly formal labels that seem constructed to be a unique field name (e.g., asking for the satisfaction score composite or the on-air-duration).\\
        - Be sensitive to language that sounds like "computer-speak".\\
        - Normal domain jargon that people in that field use naturally (e.g., "earnings per share" in finance) should not be flagged.\\
    \end{addmargin}
    - Example of explicit reference: A query asking for the mean of the "EVENTTIME" column.\\
    - Example of subtle reference: A query asking for the "event message type" instead of "what kind of event happened?".\\
\end{addmargin}
- Value Reference (value\_reference: Literal["True", "Obscure", "False"]):
\begin{addmargin}[10pt]{0pt}
    - Set to "True" if: The query demonstrates clear, unambiguous knowledge of the data's content. This includes:\\
    \begin{addmargin}[10pt]{0pt}
        - Internal Identifiers: Using non-public codes, serial numbers, or IDs that are specific to a dataset (e.g., "the status of order \#A98Z-W," or "find the library book with call\_number 856.92.C4").\\
        - Unnaturally Formatted Values: Using formatting (like single or double quotes) around a value in a way that is unnatural for a typical sentence, suggesting a programmatic lookup for a value. For example, "count visitors that are 'Adult'," or 'How many people live in "Saxony".' Do not apply this rule if the quotes are used naturally, such as for a book title or a direct quote.\\
        - Value Property Descriptions: Describing the format, structure, or properties of the data values themselves, rather than just using a single value (e.g., "Find all products where the product code starts with 'SKU-' and is followed by 8 digits").\\
    \end{addmargin}
- Set to "Obscure" if: The query uses a highly specific value that is not an internal ID, but feels extremely "dataset-y" and is unlikely to be known without a table for context. This is a very narrow category for outlier queries that should only be applied to queries that contain highly specific values.\\
    \begin{addmargin}[10pt]{0pt}
        - This category does not include common named entities (people, places, movie titles), specific dates, or years. These should be classified as "False".\\
        - Example: "What is the nationality of the sprinter that finished the 100m sprint in 9.91s at the 2009 World Championship." -> Knowing the exact time is highly specific but since it comes from a public event, it is knowable and thus "Obscure".\\
    \end{addmargin}
    - Set to "False" if: The query contains no value references or if it only uses publicly knowable facts, even if they are specific. This includes:\\
        \begin{addmargin}[10pt]{0pt}
        - Common Entities: "Netherlands," "Eiffel Tower."\\
        - Specific Named Entities: "the author Haruki Murakami," "the movie 'Blade Runner 2049'."\\
        - Specific Dates \& Years: "all transactions after September 24, 2025," "results from the 2024 election."\\
        \end{addmargin}
\end{addmargin}
- Container Reference (container\_reference: bool):
\begin{addmargin}[10pt]{0pt}
    - Set to true if: The query explicitly refers to the data artifact itself. This breaks the illusion of a natural open-domain interaction by turning it into a direct instruction about a specific file or data collection.\\
    - What to look for: Look for explicit nouns that refer to the data collection, such as "dataset," "table," "file," "spreadsheet," or conceptual container specifications like "the UN survey".\\
    - Important Distinction: A reference to a part of the schema, such as a "column," "field," or "header," is a structural reference, not a container reference.\\
    - Example: "Using the provided dataset, find the top five most frequent qualifications."\\
\end{addmargin}
Query: \{query\}
\end{llmprompt}

To validate the performance and reliability of this LLM-classifier, we manually annotated a stratified sample of 145 queries. The annotation was conducted individually by two data-science experts using the same framework as the LLM. We then calculated the agreement between the two human annotators (Inter-Annotator Agreement) and the agreement between each annotator and the LLM-judge's final classification.

The results are presented in Table \ref{tab:data_independence_agreement}. Cohen's $\kappa$ scores show moderate inter-annotator agreement for structural and value references, relfecting genuine difficulty in judging whether specific terms or values constitute privileged knowledge. However, the LLM-judge achieves higher $\kappa$ with each annotator than the annotators achieve with each other across all dimensions, indicating that the classifier performs at or above human level. On the aggregate data-independence classification used in Figure~\ref{fig:data_privileged_queries}, the LLM-judge reaches substantial agreement with both annotators ($\kappa = 0.704$ and $0.600$).

The classifier shows high agreement with human experts, particularly for Structural References and Container References. While Value References had a lower inter-annotator agreement (0.676), the LLM-judge's alignment with both annotators (0.690 and 0.903) is strong, confirming its effectiveness for this task.

\begin{table}[ht]
\centering
\caption{Agreement scores for the data-independence LLM-classifier validated against two expert human annotators. Raw agreement (\%) and Cohen's $\kappa$ are reported.}
\label{tab:data_independence_agreement}
\begin{tabular}{lcccccc}
\toprule
 & \multicolumn{2}{c}{\textbf{Ann. 1 vs. Ann. 2}} & \multicolumn{2}{c}{\textbf{Ann. 1 vs. LLM}} & \multicolumn{2}{c}{\textbf{Ann. 2 vs. LLM}} \\
\cmidrule(lr){2-3} \cmidrule(lr){4-5} \cmidrule(lr){6-7}
\textbf{Dimension} & Raw & $\kappa$ & Raw & $\kappa$ & Raw & $\kappa$ \\
\midrule
Structural References & 0.869 & 0.508 & 0.903 & 0.682 & 0.897 & 0.577 \\
Value References      & 0.676 & 0.337 & 0.903 & 0.547 & 0.690 & 0.486 \\
Container References  & 0.959 & 0.728 & 0.972 & 0.851 & 0.945 & 0.664 \\
\addlinespace
Data-Independence (agg.) & 0.800 & 0.531 & 0.876 & 0.704 & 0.841 & 0.600 \\
\bottomrule
\end{tabular}
\end{table}

\FloatBarrier

\subsection{Query Ambiguity}

To classify query ambiguity, we assess each query along the five dimensions of procedural and data specification detailed in Appendix \ref{sec:appendix:query_specification_framework}. This classification determines whether a query is \textbf{unambiguous}, as defined in Section 2 and analyzed in Figure \ref{fig:query_specification}.

We developed two distinct LLM-based classifiers: one for Data Specification (Prompt \ref{prompt:data_spec_prompt}) and one for Procedural Specification (Prompt \ref{prompt:procedural_spec_prompt}). Both classifiers were run using gpt-5-2025-08-07 on the same 500-query samples from each dataset.

The classifiers provide nuanced labels for each of the five dimensions introduced in Appendix \ref{sec:appendix:query_specification_framework}. For the final analysis, we aggregate these labels into a boolean flag indicating if a dimension is sufficiently specified (True) or not (False). A query is considered specified for a dimension if it meets the following criteria: \begin{itemize} \item \textbf{Entities:} Classified as "Specified". \item \textbf{Temporal:} Classified as "Specified", "Underspecified (Assuming Recency)", or "Not Applicable". \item \textbf{Domain:} Classified as "Specified" or "Underspecified (Assuming Universal Domain)". \item \textbf{Intent:} Classified as "Specified". \item \textbf{Methodological:} Classified as "Specified". \end{itemize}

We then derive the higher-level categories used in our analysis by combining these boolean flags: \begin{itemize} \item \textbf{Data Specification} = Entities AND Temporal AND Domain \item \textbf{Procedural Specification} = Intent AND Methodological \item \textbf{Unambiguous} = Data Specification AND Procedural Specification \end{itemize}

To verify the efficacy of this classification approach, we tasked a human data-science expert with reviewing and correcting the LLM-generated labels for a random sample stratified over the datasets of 148 queries. 
We chose this correction-based methodology because the classification task requires broad (world) knowledge to assess whether query components can be resolved through common-sense or convention. In initial tests, disagreements between annotators frequently stemmed from differences in domain knowledge rather than judgment, making a correction-based approach where the expert reviews and overrides only genuine classification errors more appropriate than independent dual annotation. The results in Table~\ref{tab:ambiguity_agreement} show strong agreement across all dimensions, with $\kappa \geq 0.73$ on every individual dimension and $\kappa = 0.866$ on the aggregate unambiguous classification used in Figure~\ref{fig:query_specification}.

\begin{table}[htbp]
    \centering
    \caption{Agreement scores for the query ambiguity LLM-classifiers, validated against expert-corrected labels. Raw agreement (\%) and Cohen's $\kappa$ are reported.}
    \label{tab:ambiguity_agreement}
    \begin{tabular}{lcc}
        \toprule
        \textbf{Specification Level} & \textbf{Raw} & $\boldsymbol{\kappa}$ \\
        \midrule
        Entities & 0.953 & 0.895 \\
        Temporal & 0.887 & 0.732 \\
        Domain & 0.960 & 0.911 \\
        Intent & 1.000 & 1.000 \\
        Methodology & 0.927 & 0.811 \\
        \addlinespace
        Data (agg.) & 0.947 & 0.832 \\
        Procedure (agg.) & 0.933 & 0.862 \\
        \addlinespace
        Unambiguous (agg.) & 0.953 & 0.866 \\
        \bottomrule
    \end{tabular}
\end{table}

\begin{llmprompt}{Data-Specification Prompt}
\label{prompt:data_spec_prompt}
\tiny

You are an expert data analyst and annotator. Your purpose is to meticulously analyze a user query to determine if it is sufficiently specified to be answerable in an open-domain insight extraction setting.
\\
An "open-domain" system must identify relevant data from a massive, unknown corpus of tables before executing a query. For this to be possible, the query must be self-contained and unambiguous, providing enough detail to pinpoint the correct data without needing to ask for clarification.
\\
Core Principle:
\\
Before you begin, apply this mental model: Imagine you must give the query to a researcher who has access to every table in the world but cannot ask you any clarifying questions. Your task is to decide if they could find the a relevant set of tables to work on. If they would have to ask clarifying questions like "which employees?" or "for what year?", the query's data specification is flawed.
\\
\\
Data Specification Dimensions:
\\
You will analyze the query based on the following three dimensions of data specification.
\\
1. Entity Specification (Binary: {Specified} / {Underspecified}) (Clarifying questions are typically: "What...?", "Which...?", "Whose...?", "Who...?")
\\
This dimension evaluates the core subjects (nouns) of the query based on the Principle of Reasonable Resolvability: an entity is specified if a capable system can be expected to find and disambiguate it with high confidence using the query's context and world knowledge. The classification is based on a consideration of all entities in the query highlighting any that are underspecified. Focus on ambiguity regarding the data entities, not the actions or metrics.
\\
\vspace{-0.2cm}
\begin{addmargin}[10pt]{0pt}
* Specified: The entities in the query are specific enough to be confidently identified using the contents of the query and world knowledge. There is two main ways this can be true:
\begin{addmargin}[10pt]{0pt}
1. Unique identification: The query refers to one or more specific entities that are uniquely named or can be confidently resolved to a single instance (e.g., "the 2024 Tour de France," "GDP of Brazil," "Ronaldo" (in the context of football)). This includes entities that require multi-step lookups or contextual disambiguation (e.g., "the actor who played the main character in the 2005 film 'The Great Film'" is specified if that film is unique).
\\
2. Broad classes: The query refers to a well-defined class of entities, where the user's intent is to query over the entire class. This also includes concepts where finding a suitable dataset is part of the challenge. Broad classes are also specified in cases where "any" is a reasonable interpretation (e.g., "Who was the first female to attend a german university?" -> here "german university" is a broad class, but "any german university" is a reasonable and specified interpretation within the query).
\end{addmargin}
* Underspecified: The query refers to an entity that is truly ambiguous even with world knowledge, where multiple plausible candidates exist and the query provides no path to resolution (e.g., "the recent race," "the company's revenue").
\\
\\
Note: The fact that broad classes of entities may yield large or complex datasets does not make them underspecified, i.e. just because a class of entities can be specified further, does not mean that this specification is necessary to resolve ambiguity. For example, "all employees" is specified if the query is about a specific company, even if that company has many employees and subsidiaries.
\end{addmargin}
2. Temporal Specification (Quaternary: {Specified} / {Underspecified (Assuming Recency)} / {Underspecified (Ambiguous)} / {Not Applicable}) (Clarifying questions are typically: "When...?", "For what time period...?", "As of when...?")
\\
\\
This dimension evaluates the time frame used to filter the data.
\\
\vspace{-0.2cm}
\begin{addmargin}[10pt]{0pt}
* Specified: The query provides a clear time frame, which can be an explicit range ("in 2023"), a resolvable term ("last year"), or an implicit "all-time" scope assumed for superlative/cumulative queries ("Who has won the most Oscars?"). Use this classification for queries that are implicitly applicable in an "all-time" context, such as superlative or cumulative queries (e.g., "largest economy," "most decorated Olympian").
\\
* {Underspecified (Assuming Recency)}: The query lacks a time frame, but the probable intent is the most recent available data.
\begin{addmargin}[10pt]{0pt}
* 'Assuming Recency' should be considered the standard default for general statistical queries about current states, such as populations, demographics, economic indicators, or inventory counts (e.g., a query about "the number of hospitals" or "the unemployment rate").
\\
* Use 'Assuming Recency' for queries about data that is periodically updated or versioned, where a common-sense intent is to use the latest available data (e.g., population statistics, economic indicators, inventory counts).
\end{addmargin}
* Underspecified (Ambiguous): The query lacks a time frame, and the intent is truly ambiguous with no safe default, i.e., it requires a time frame or timing to make sense ("What was the stock price of Apple?").
\\
* {Not Applicable}: The query is independent of a specific time or time-frame and concerns a stable, definitional fact that is not expected to change, such as (historical) events, physical constants, or mathematical definitions (e.g., "When was the Eiffel Tower completed?"). Do not use this for data that is versioned or updated, even if infrequently.
\\
\\
Note: Queries that get their temporal specification by making reference to the data container (e.g., "in the dataset," "in the table") are classified as "Underspecified (Ambiguous)" since the data container is unknown in an open-domain setting.
\\
\\
Note: Recency can often be assumed in queries formulated in the present tense (e.g., "What is the population of...?" or "How many employees does... have?"). However, if a query is clearly about a past event or state but lacks a time frame (e.g., "What was the population of...?" or "How many employees did... have?"), it should be classified as {Underspecified (Ambiguous)}.
\end{addmargin}
3. Domain Specification (Ternary: {Specified} / Underspecified (Assuming Universal Domain) / Underspecified (Ambiguous) ) (Clarifying questions are typically: "Where...?", "In which/whose/what...?")
\\
This dimension evaluates the contextual boundary (geographical, organizational, conceptual) used to filter the data.
\\
\vspace{-0.2cm}
\begin{addmargin}[10pt]{0pt}
* Specified: The query defines a boundary that is globally unique or self-contained in its common-sense context. Also use this category if the query's contextual boundary is inherently unambiguous due to common knowledge or the query's concept has no inherent domain boundary ("What is the element with the atomic number of 1?").
\begin{addmargin}[10pt]{0pt}
* This includes globally unique entities (e.g., "companies in Japan," "clubs in the English Premier League").
\\
* This also includes entities with a single, dominant interpretation (e.g., a query about "the Bay Area" clearly implies the San Francisco Bay Area, "Ronaldo" likely refers to the football player "Christiano Ronaldo").
\end{addmargin}
* Underspecified (Assuming Universal Domain): The query lacks a specified domain boundary, but has a sensible, widest-possible default interpretation (e.g., "highest mountain" -> world, "largest economy" -> global).
\\
* Underspecified (Ambiguous): The query lacks a boundary, and the context is truly missing ("highest-paid employees," which requires a company). Only use this category if the query is nonsensical without a specific domain boundary.
\\
\\
Note: Queries that get their domain boundary specification by making reference to the data container (e.g., "in the dataset," "in the table") are classified as "Underspecified (Ambiguous)" since the data container is unknown in an open-domain setting.\\
\\
Query: {query}
\end{addmargin}
\end{llmprompt}

\begin{llmprompt}{Procedural Specification Analysis}
\label{prompt:procedural_spec_prompt}
\tiny
You are an expert data analyst and annotator. Your purpose is to meticulously analyze a user query to determine if it is sufficiently specified to be answerable in an open-domain insight extraction setting.

For this to be possible, the query must be self-contained and unambiguous, providing enough detail to execute an analytical procedure without requiring further user input.

Core Principle:
Imagine an analyst has already located the perfect, unambiguous dataset for the query (e.g., they have the table of "all player statistics for the 2024-2025 English Premier League season"). Your job is to decide if the query gives them clear enough instructions on what to do with that data. If they would have to come back and ask you, "how should I calculate 'top'?" or "what do you mean by 'relationship'?", then the query is procedurally underspecified.

For this assessment, you must assume the data has already been perfectly identified. Do not worry about whether entities (like "players"), timeframes, or domains are ambiguous. Your only job is to assess the clarity of the analytical steps requested. Focus only on the "what to do and how to do it," not the "what data."

Procedural Specification Dimensions:

You will analyze how the query specifies the analytical operations the user wants to perform on the data.

1. Intent Specification (Binary: Specified / Underspecified) (Clarifying questions are typically: "What does ... mean?",...)

This dimension evaluates the overall analytical action requested. (i.e., the user's goal)
\begin{itemize}
    \item Specified: The query clearly states an executable operation (e.g., list, count, rank, calculate the average, find the correlation).
    \item Underspecified: The query's intent is vague (e.g., "tell me about," "information on", "relationship between").
\end{itemize}

2. Methodological Specification (Binary: Specified / Underspecified) (Clarifying questions are typically: "How to...?", "By what metric...?",...)

This dimension evaluates the clarity of the calculations and transformations applied to the data (i.e., the method to achieve the goal).

If ambiguity arises from which data column or entity to use (e.g., for filtering the data), this is a Data Specification issue (Part B), not a Scope Specification issue.
\begin{itemize}
    \item Specified: The query does not require any calculations or analytical methods, or these methods are sufficiently specified:
    \begin{itemize}
        \item The query is a simple lookup that does not require complex analytical parameters (e.g., "List the names of drivers...", "What role did she play?").
        \item The query requires some calculation or application of an analytical method, and relevant parameters for that calculation are clear or can be assumed by a standard, common-sense default (e.g., using a standard unweighted average, using a logical (time) unit). The explicit criteria include:
        \begin{itemize}
            \item Ranking: The specific metric used for ordering (e.g., top 5 by revenue).
            \item Grouping \& Aggregation: The categories for grouping and the function to apply (e.g., the average salary per department).
            \item Metric Calculation: The formula for any derived values (e.g., GDP per capita).
            \item Analytical Model: The method for assessing a relationship (e.g., the correlation between...).
        \end{itemize}
    \end{itemize}
    \item Underspecified: The query is missing an important analytical parameter. Use this category only if a missing parameter represents a significant analytical choice that could materially change the query's intent or result (e.g., the metric for "best," the regression method to use). Do not use it for instances where a reasonable default can be assumed without clarification.
\end{itemize}

Query: \{query\}
\end{llmprompt}

\end{document}